\documentclass[letterpaper, 10 pt, conference]{ieeeconf}
\IEEEoverridecommandlockouts
\usepackage{comment}
\usepackage{cite}
\usepackage{booktabs}
\usepackage{array} 
\usepackage{caption} 
\usepackage{amsmath,amssymb,amsfonts}
\usepackage{algorithmic}
\usepackage{graphicx}
\usepackage{textcomp}
\usepackage{xcolor}
\usepackage{subcaption}
\usepackage{multirow}
\usepackage[normalem]{ulem}
\useunder{\uline}{\ul}{}
\usepackage{cancel}

\def\BibTeX{{\rm B\kern-.05em{\sc i\kern-.025em b}\kern-.08em
T\kern-.1667em\lower.7ex\hbox{E}\kern-.125emX}}

\begin{document}

\title{\LARGE \bf
Motivating Students' Self-study \\with Goal Reminder and Emotional Support\\
}

\author{
    Hyung Chan Cho$^{*1}$, Go-Eum Cha$^{1}$, Yanfu Liu$^{1}$, Sooyeon Jeong$^{1}$ 
    \thanks{$^{*}$ Corresponding author: {\tt\small cho234@purdue.edu}}
    \thanks{$^{1}$ Department of Computer Science, Purdue University, West Lafayette, IN 47906, USA}
}

\maketitle

\begin{abstract}
While the efficacy of social robots in supporting people in learning tasks has been extensively investigated, their potential impact in assisting students in self-studying contexts has not been investigated much. This study explores how a social robot can act as a peer study companion for college students during self-study tasks by delivering task-oriented \textit{goal reminder} and positive \textit{emotional support}. We conducted an exploratory Wizard-of-Oz study to explore how these robotic support behaviors impacted students' perceived focus, productivity, and engagement in comparison to a robot that only provided physical presence (\textit{control}). Our study results suggest that participants in the \textit{goal reminder} and the \textit{emotional support} conditions reported greater ease of use, with the \textit{goal reminder} condition additionally showing a higher willingness to use the robot in future study sessions. Participants' satisfaction with the robot was correlated with their perception of the robot as a social other, and this perception was found to be a predictor for their level of goal achievement in the self-study task. These findings highlight the potential of socially assistive robots to support self-study through both functional and emotional engagement.
\end{abstract}


\section{Introduction}
Peer relationships in educational settings play a crucial role in generating relatedness and support that are influential in fostering academic success \cite{lavy2017benefits, hockings2018independent, tikkanen2024peer, kosonen2024university}. Peer support is shown to positively impact students’ learning by fostering a sense of connectedness, which enhances productivity, academic performance, and study well-being \cite{lavy2017benefits, byrom2018evaluation, altermatt2019academic, tikkanen2024peer}. Peer-based settings have also been effective in addressing procrastination, as students engage with one another to stay motivated and focused on the task \cite{cushing1997academic, milem1998attitude, carter2005effects}. Shared motivation and emotional support are key to sustaining academic progress in these contexts \cite{kindermann1993natural, cushing1997academic, milem1998attitude, carter2005effects}.
Although the benefits of peer support are well recognized, some students experience barriers in engaging in a peer study group, such as social anxiety, fear of judgment \cite{downing2020fear, k2021meeting}, mismatched expectations \cite{barfield2003students}, and dynamic group conditions \cite{dokuka2015diffusion}, which often hinder sustained peer interaction and lead them to study alone, i.e., self-study.

Self-study requires self-regulation, and students employ various strategies to achieve their academic goals, e.g., breaking down goals, minimizing distractions, or utilizing time management techniques. Yet, many college students often struggle to stay focused and motivated over an extended period \cite{zimmerman2002becoming, shalev2018use} and fail to self-regulate due to distraction that provides instant rewards (e.g., social media, digital entertainment, etc.) or negative emotions associated with the study task \cite{muraven2000self, shalev2018use}. 

We envision that social robots (SRs) have unique opportunities to address these gaps as a study companion in college students' self-study context due to their ability to provide socio-emotional support and accountability. So far, SRs have primarily been explored in educational context as a teacher/tutor \cite{park2011teaching} or as a peer-learner \cite{dias2008sliding} that can deliver both educational content and socio-emotional support during learning tasks \cite{leite2012modelling,kennedy2015higher, short2014train,leyzberg2012physical,saerbeck2010expressive, jeong2020robotic, jeong2023deploying,o2024design}. However, there is little work that explores how SRs can provide self-regulatory and emotional support to students during a self-study task context. 

Thus, we present a robotic companion that acts as a supportive peer and provides motivational support by (1) presenting peer-like behavior only, (2) reminding students about their task goals (\textit{goal reminder}) or (3) providing positive emotional support (\textit{emotional support}) (Fig.~\ref{fig:experimental_conditions}). We conducted a between-subject study with 76 participants to explore students' perceptions toward the robot's role as a study companion and their overall experience with the robot interventions. Particularly, we examined how the robot's suggestions to re-focus and positive encouragement influenced their perceived concentration and productivity, and how their task behavior changed after each of the robot's motivational supports.


\section{Related Works}

\subsection{Self-regulation Strategies for Self-study }
Self-regulation is essential for effective self-study, enabling students to set goals, manage time, and sustain focus \cite{virtanen2015self}. Key strategies include goal-setting, structured study sessions, and minimizing distractions, but maintaining self-regulation over extended periods can be challenging due to procrastination and susceptibility to distractions \cite{zimmerman2002becoming, muraven2000self, shalev2018use}. 
External support, such as peer interaction, can help by offering motivational support, accountability, and emotional reinforcement \cite{cushing1997academic, milem1998attitude, carter2005effects}. Study conditions or co-working sessions foster shared responsibility and resilience, highlighting the role of social engagement in reinforcing self-regulation \cite{dang2023students, didonato2013effective}. This suggests that external support, whether from peers or other study companions, can enhance students' ability to maintain focus and achieve their learning objectives \cite{kindermann1993natural, cushing1997academic}. Beyond human peers, technology also aids self-regulation through structured techniques and digital tools. The Pomodoro Technique \cite{cirillo2018pomodoro}, a time-management strategy that segments study sessions into focused intervals with breaks, has been widely adopted by various digital tools to enhance concentration and reduce cognitive fatigue. Distraction-limiting features, such as the \textit{Do Not Disturb} mode on smartphones, further support sustained focus \cite{pielot2017productive}. These approaches, whether through social reinforcement or technological interventions, help students sustain concentration, manage study habits, and improve productivity.

\subsection{Technologies for Self-Regulation and Productivity}
Various technological tools have been developed to support self-regulation and enhance productivity by reinforcing goal-setting, minimizing distractions, and maintaining focus. For instance,  StudyBuddy\footnote{https://www.studybuddymobile.com/} integrates peer-based accountability with self-regulation techniques, enabling users to set study goals and participate in virtual study sessions that simulate the presence of a peer. 
Forest: Stay Focused\footnote{https://www.forestapp.cc/}  incentivizes users to avoid using other applications on their phones by growing virtual trees when they concentrate on their tasks.
While these technologies provide external support for maintaining focus, they often lack the embodied presence and social interaction that are essential for motivation and sustained engagement \cite{yang2006exploring,thoman2007talking}, which are present in peer-based study groups. 

In comparison to other technological mediums, social robots could not only provide self-regulation support (e.g., reminders to re-focus on the task when distracted) but also peer-like companionship support. For instance, O'Connell et al. \cite{o2024design} developed a SR study companion for college students with ADHD and many participants perceived the robot positively and chose to continue using it after one-week interaction. Although applied in a different application context, Kory and Breazeal \cite{kidd2008robots} found that a physically embodied robot prompted greater behavior change and sustained engagement than a computer system or a paper-based intervention in a long-term weight management support context.  
These findings suggest that SRs could bridge the gap between digital interventions and human peer support, making them a promising solution to support students' study habits and engagement.

\subsection{Social Robots for Supporting Students}
Social robots have been increasingly integrated into educational settings to enhance learning experiences and support student outcomes \cite{greczek2014socially}. Their interactive and adaptive behaviors have been shown to foster motivation, improve academic performance, and support cognitive development \cite{belpaeme2018social}. Prior research has demonstrated their effectiveness in various educational contexts, such as improving cognitive task performance \cite{leyzberg2014personalizing}, 
and adapting instructional strategies to enhance engagement \cite{schadenberg2017personalising}. Additionally, studies have explored how social robots influence learning by modulating expressive characteristics \cite{kory2017flat}, optimizing study session timing \cite{ramachandran2017give}, and responding to learners' emotional states \cite{gordon2016affective}. Another study by Jeong et al. \cite{jeong2023deploying, jeong2020robotic} found that their robotic positive psychology coach was effective in improving college students' psychological well-being, mood, and readiness for behavioral change, with user-robot rapport and personality traits influencing intervention effectiveness. Despite these promising findings, there remains a gap in understanding the types of supportive interventions and behaviors social robots exhibit to better support students' self-study as physically embodied study companions. In this work, we explore the task-related and affect-related strategies for robotic study companions to support students' motivation, engagement, and productivity in self-study tasks.

\section{The Exploratory Wizard-or-Oz Study}
\subsection{Participants}
We recruited $N$=$76$ students who were (1) 18 years or older, (2) fluent in English, and (3) regularly engaged in academic tasks at Purdue University. Of the seventy-six participants, two participants were excluded from the analysis; one participant voluntarily withdrew from the study, and the other participant failed to comply with the study protocols. 74 participants ($M$=$21.04$, $SD$=$3.60$, range=$18–33$; 32 female, 40 male, 1 prefer not to say, 1 other) completed the full study. There were 19 freshmen, 18 sophomores, 12 juniors, 8 seniors, 16 graduate students, and 1 other. A block randomization design was used to balance participants' age and gender across the experimental conditions. Twenty-one participants were assigned to the \textit{control} condition, twenty-seven participants were assigned to the \textit{goal reminder} condition, and twenty-six participants were assigned to the \textit{emotional support} condition. Our study was approved by Purdue University's Institutional Review Board (IRB) (\#IRB-2024-608).

\subsection{Study Procedure} \label{method}
Upon arrival at the research site, participants were briefed on the overall study procedure and signed the written informed consent form to participate in the study. Then, they were asked to complete a set of pre-study questionnaires to collect information about their demographic and usual self-study behaviors and habits (Section \ref{sec:self-reported}).

\begin{figure*}[ht]
    \centering
    \includegraphics[width=0.95\linewidth]{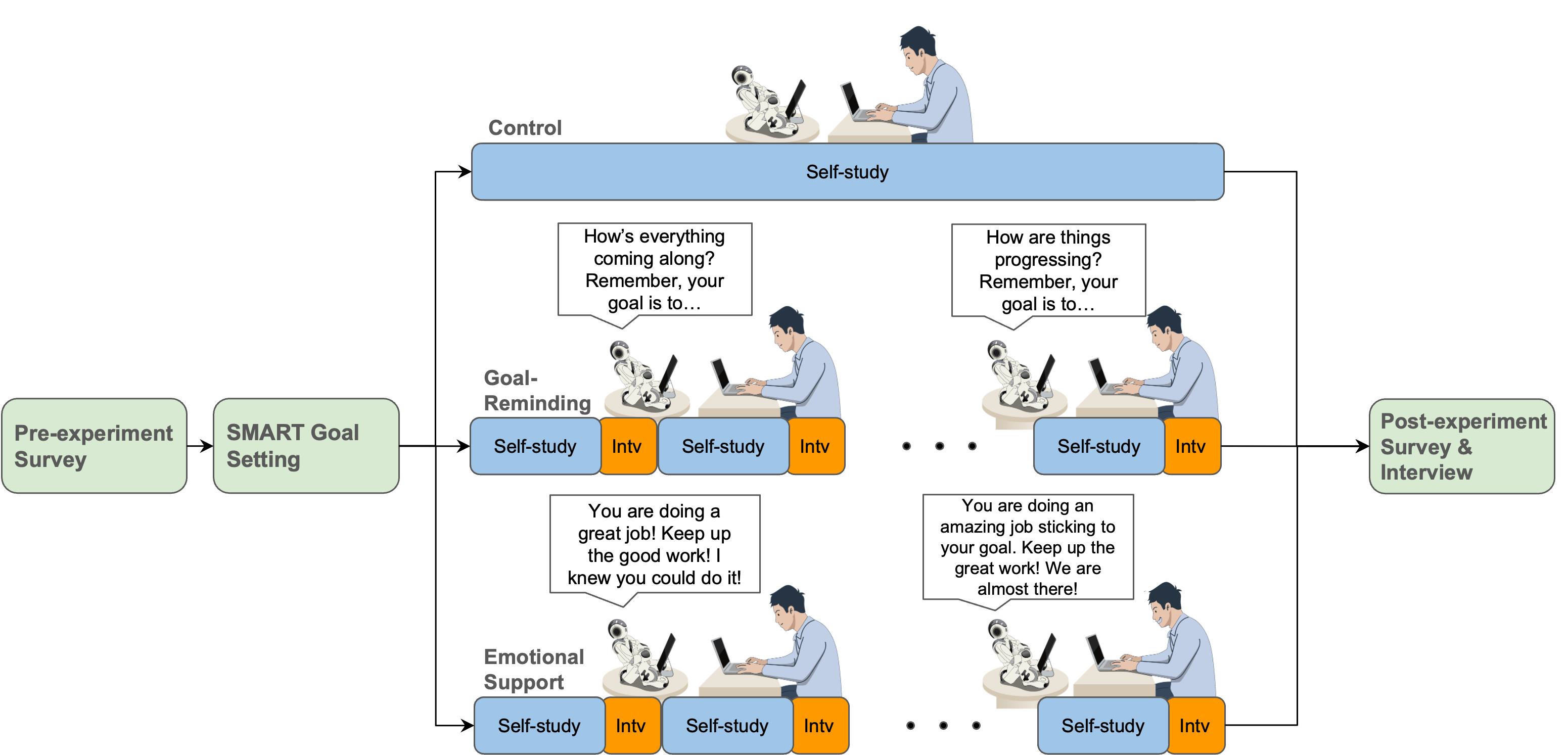}
    \caption{Participants were randomized into one of the three experimental conditions. In the \textit{control} condition, the robot answered questions from students without initiating any intervention during the study session. In the \textit{goal reminder} condition, the robot reminded the participant of their self-study goal when they looked distracted or not focused. In the \textit{emotional support} condition, the robot provided encouraging and positive words to motivate the participant to stay focused and to keep working on the self-study task.}
    \label{fig:experimental_conditions}
    \vspace{-12pt}
\end{figure*}

\begin{figure}
    \centering
    \includegraphics[width=\linewidth]{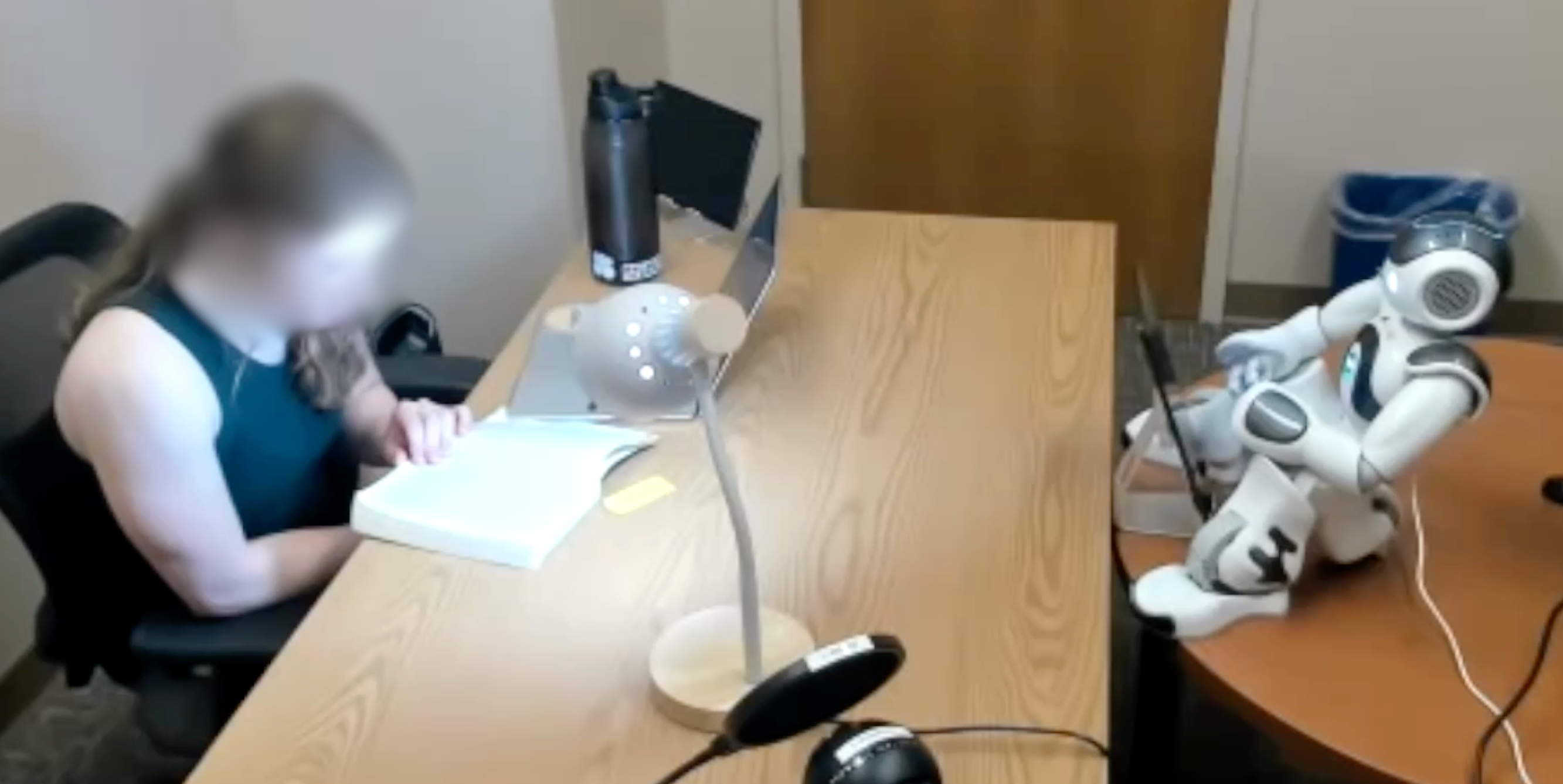}
    \caption{The robot and the participant engaged in self-studies for two hours.}
    \label{fig:study_setup}
    \vspace{-12pt}
\end{figure}
\raggedbottom

Our study was conducted in a quiet office space where the participants could engage in self-study for two hours with a NAO robot. To capture students' realistic study behaviors, participants were informed to bring their own study materials (e.g., homework assignment) prior to our study. While this introduced variability in study content, we chose to have students choose their own study/work materials for more authentic and natural self-study behaviors. Each participant interacted with the NAO robot that acted as a study companion. First, the robot guided the participant to set a SMART goal (Specific, Measurable, Achievable, Relevant, Time-bound) \cite{doran1981there} for the two-hour self-study session (e.g., ``I want to solve three LeetCode problems''). Once the participant set their goal, both the robot and the participant started their self-study activities. The robot ``read'' a digital book on an Android tablet and occasionally turned the virtual page with a flipping arm gesture (Fig. \ref{fig:study_setup}). When participants asked what the robot was reading, the robot responded briefly and redirected the participant back to their task to maintain the study flow.

During this self-study phase, the robot exhibited three different behaviors depending on the experimental condition. In the \textit{control} condition, the robot did not intervene with the participant until the end of the study session and kept reading its digital book.
In the \textit{goal-reminder} condition, the robot reminded the participant's SMART goal (e.g., ``how's everything coming along? Remember, your goal is to [complete the problem set.]''). In the \textit{emotional support} condition, the robot provided emotionally supportive statements to encourage the participants' progress so far (e.g., ``You're making great progress on your reading!''; see Fig.~\ref{fig:experimental_conditions})

All robot behaviors were remotely controlled, and each robot intervention in the \textit{goal reminder} and the \textit{emotional support} conditions was performed when participant showed signs of distraction, e.g., dozing off, using their mobile devices, appearing frustrated, yawning, stretching, doodling, or looking around the room in distraction. If none of these behaviors were observed, an intervention was given after 30 minutes from the last intervention or the beginning of the study session. This approach ensured the consistency of both the timing and content of interventions, not disturbing participants frequently. Across the study, each session included between four to six interventions based on the participant's progress and the timing of the session. After each intervention, the NAO robot asked the participants to complete a short survey to provide their feedback on the overall satisfaction, engagement, and timing of the intervention (Fig.~\ref{fig:ema}).

After the self-study session, participants completed the post-study questionnaire (Section \ref{sec:self-reported}) and engaged in a semi-structured interview (Section \ref{sec:data_interview}) to provide qualitative feedback on the overall study experience and our robot study companion. 
The interviews were audio-recorded for transcriptions and data analysis.

Participants were compensated based on the activities completed: \$5 each for pre- and post-surveys, \$10 for an interview, and \$20 for completing in-session surveys. Thus, participants in the \textit{goal reminder} and \textit{emotional support} conditions received \$40, and those in the \textit{control} condition received \$20. To reflect this difference, recruitment occurred in two separate phases with condition-specific consent forms.

\subsection{Data Collection}
\subsubsection{Self-reported Measures}
\label{sec:self-reported}
Prior to the study, participants completed a pre-study questionnaire on their demographic information, baseline study habits, their perceived stress/concentration on academic workload, and current strategies for focus using the Attitude Toward Technology in Teaching and Learning (ATTC) scale \cite{derryberry2002anxiety,attentioncontrolscale}.  

\begin{figure}[t]
    \centering
    \includegraphics[width=\linewidth]{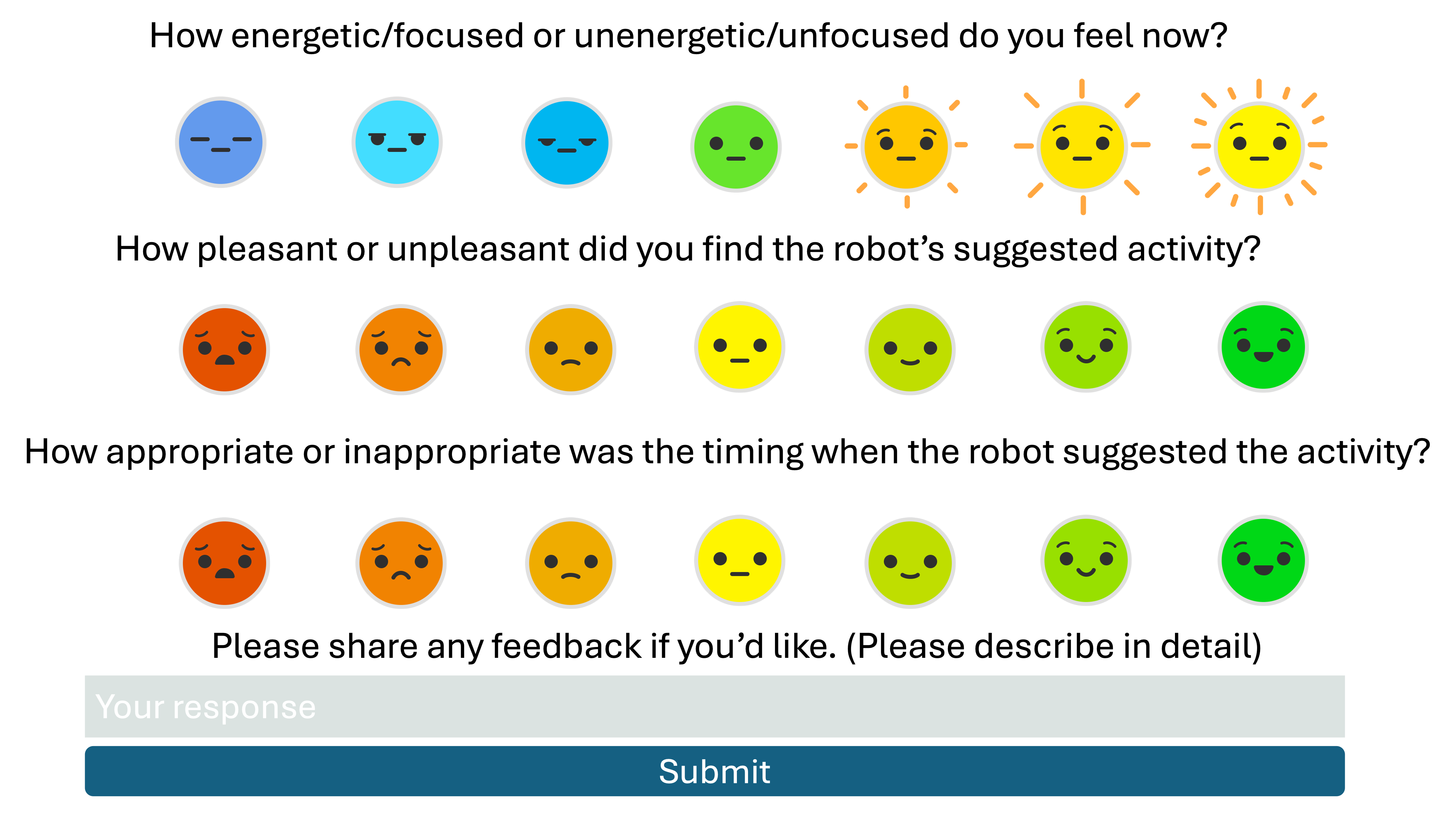}
    \caption{Participants completed a quick picture-based evaluation of the robot's motivational interventions.}
    \vspace{-8pt}
    \label{fig:ema}
\end{figure}

During the self-study session with the robot, participants in the \textit{goal reminder} and the \textit{emotional support} conditions were asked to fill out a short picture-based survey immediately after each robot intervention (Fig.~\ref{fig:ema}). The three questions measured participants' affective states and the appropriateness of timing for the robot intervention. We also included an optional open-ended text response in case participants wanted to provide any other feedback. 

\raggedbottom

After the self-study session, participants completed a post-study questionnaire to measure their perceived productivity (PP) and concentration level (PC) during the self-study session, effectiveness of the robot study companion, and the acceptance of assistive robot scale developed by Heerink et al. \cite{heerink2009measuring}, which measured the effectiveness of robot intervention (ERI), the robot's perceived ease of use (PEOU), perceived usefulness (PU), perceived enjoyment (PENJ), trust in the robot (TRUST), and overall satisfaction \cite{heerink2009measuring}.

\subsubsection{Post-study Interview}
\label{sec:data_interview}
At the end of the study, we also interviewed participants to obtain qualitative feedback on their perceived effectiveness of our robot study companions and the overall study procedure. The interview included questions regarding participants' overall experience,
the perceived helpfulness of the robot (e.g., ``Were there any specific interactions or behaviors of the robot that you found particularly helpful or unhelpful?''), and the effect of the robot’s presence in their study habits, mood, or stress levels (e.g., ``How did the robot's presence and interventions affect your mood or stress levels during the study session?''). Participants were also asked to provide feedback on the robot’s features, interaction frequency, and potential future improvements. 

\subsubsection{Video and Audio Data}
Participants' study sessions with the robot and their semi-structured interviews were video/audio recorded with consent. The audio data were transcribed using the Google Speech-to-Text API, with manual correction applied to the generated transcripts.

\section{Data Analysis} 
\label{sec:data_analysis}

\subsection{Statistical Analyses for Self-report Measures}
Participants' post-questionnaire responses were not normally distributed according to the Shapiro-Wilk test and thus we conducted non-parametric tests for our data analyses. We used Kruskal-Wallis tests with post-hoc Dunn's tests to compare participants' responses to post-study surveys and behavioral outcomes across the three experimental conditions. Mann-Whitney U tests were used to compare study outcomes only available in the \textit{goal-reminder} and the \textit{emotion support} conditions. For multiple comparisons, we applied the Benjamini-Hochberg corrections to control for the false discovery rate. Chi-square tests were utilized to run independence between data from experimental conditions and goal-achievement levels. To test associations between participants' perception on the robot and their responses, we also conducted Spearman's rank correlation tests. Ordinal regression analysis was performed to see the predictive power on goal-achievement levels based on maximum likelihood estimation. The fitness of the regression was estimated with Cox and Snell's pseudo $R^2$ \cite{cox2018analysis}. 

\subsection{Behavioral Annotations and Analyses}
We annotated participants' behaviors observed during the study sessions based on the recorded interaction footage and the transcription data to investigate how each robotic intervention influenced their self-study behavior and attitude toward the robot study companion. 
We manually annotated the following participant behaviors: (1) the level of the SMART goal achievement, (2) participants' human-like perception of the robot, and (3) overall explicit affect/emotion observed throughout the study session. 

We set participants' self-selected SMART goals co-created with the NAO robot as the target goal and categorized each participant into one of the three achiever types: \textit{under-achiever}, those who did not complete the self-study goal; \textit{exact-achiever}, those who completed the exact amount of their goal; and \textit{over-achiever}, those who completed their initial goal and conducted additional self-study tasks. We used one of the post-study survey items (``Did you achieve the short-term study goal you set at the beginning of the session?") to categorize participants into one of the three distinct conditions. However, as the survey item allowed answers of ``yes," ``no," and ``partially," we reviewed the interaction transcription to verify their confirmatory utterances during the experiment. For example, participants who completed their original SMART goals and set an additional goal during the study session were categorized as over-achievers. 

We also annotated participants' self-study behavior observed immediately before and after the robot interventions to evaluate how the robot's supportive behavior influences students' focus and attention on the self-study task. We observed the interaction recording at two time points (30 seconds before and after each robot intervention) and annotated participants'  behaviors as either \textit{focused} or \textit{distracted}. This annotation was only conducted for the \textit{goal reminder} and the \textit{emotional support} conditions as the robot did not provide any intervention in the \textit{control} condition. 

In order to investigate participants' perceptions of the robot as a study companion rather than a productivity tool, we annotated participant behaviors that suggested that they perceived the NAO robot as a ``social other.'' Examples of such behaviors include asking the robot what it was reading, exploring the robot's capabilities, and using verbal (e.g., ``sounds good'' or ``thank you") or non-verbal social communications (e.g., showing two thumbs) to the robot. We specifically annotated these behaviors at two phases of the study session: (1) during the SMART goal-setting interaction and (2) during the wrap-up phase when the robot notified the end of the self-study session. Two independent annotators coded these behaviors and achieved a substantial inter-rater reliability level based on Cohen's Kappa $\kappa$=0.702. 

\subsection{Qualitative Analysis on Post-study Interview Data}
We conducted a thematic analysis \cite{clarke2017thematic} on the post-study interview data to investigate participants' engagement, focus, and motivation during the study sessions, and their attitudes toward the robot study companion. Two independent researchers extracted preliminary and sub-themes from the transcripts. The final themes were consolidated through in-depth discussion among the research team members. 

\section{Results}

\subsection{Students' Baseline Focus and Attention }
Based on the pre-study questionnaire responses, we found that our study participants typically experience moderate levels of perceived workload ($M$=$4.27$, $SD$=$1.734$, range=1-7) and concentration quality ($M$=$4.243$, $SD$=$1.392$, range=1-7). 
Participants also reported moderate levels of difficulties in maintaining focus when exposed to various distraction factors: events ($M$=$2.202$, $SD$=$0.684$, range=1-4), internal thoughts ($M$=$2.364$, $SD$=$0.7$, range=1-4), and regulating excitement ($M$=$2.506$, $SD$=$0.893$, range=1-4).
They reported a moderate level of capacity in managing multiple tasks, such as task switching one to another ($M$=$2.756$, $SD$=$0.57$, range=1-4), delay in a new task ($M$=$2.432$, $SD$=$0.604$, range=1-4), attentional shift to a new topic ($M$=$2.91$, $SD$=$0.65$, range=1-4), and alternating between two tasks ($M$=$2.554$, $SD$=$0.716$, range=1-4). Participants indicated limited ability to refocus on ongoing tasks after experiencing distractions ($M$=$2.486$, $SD$=$0.801$, range=1-4).
\raggedbottom

\subsection{Self-reported Perceptions on Each Robot Intervention} \label{section:account}
\begin{figure}[t]
    \centering
    \includegraphics[width=0.98\linewidth]{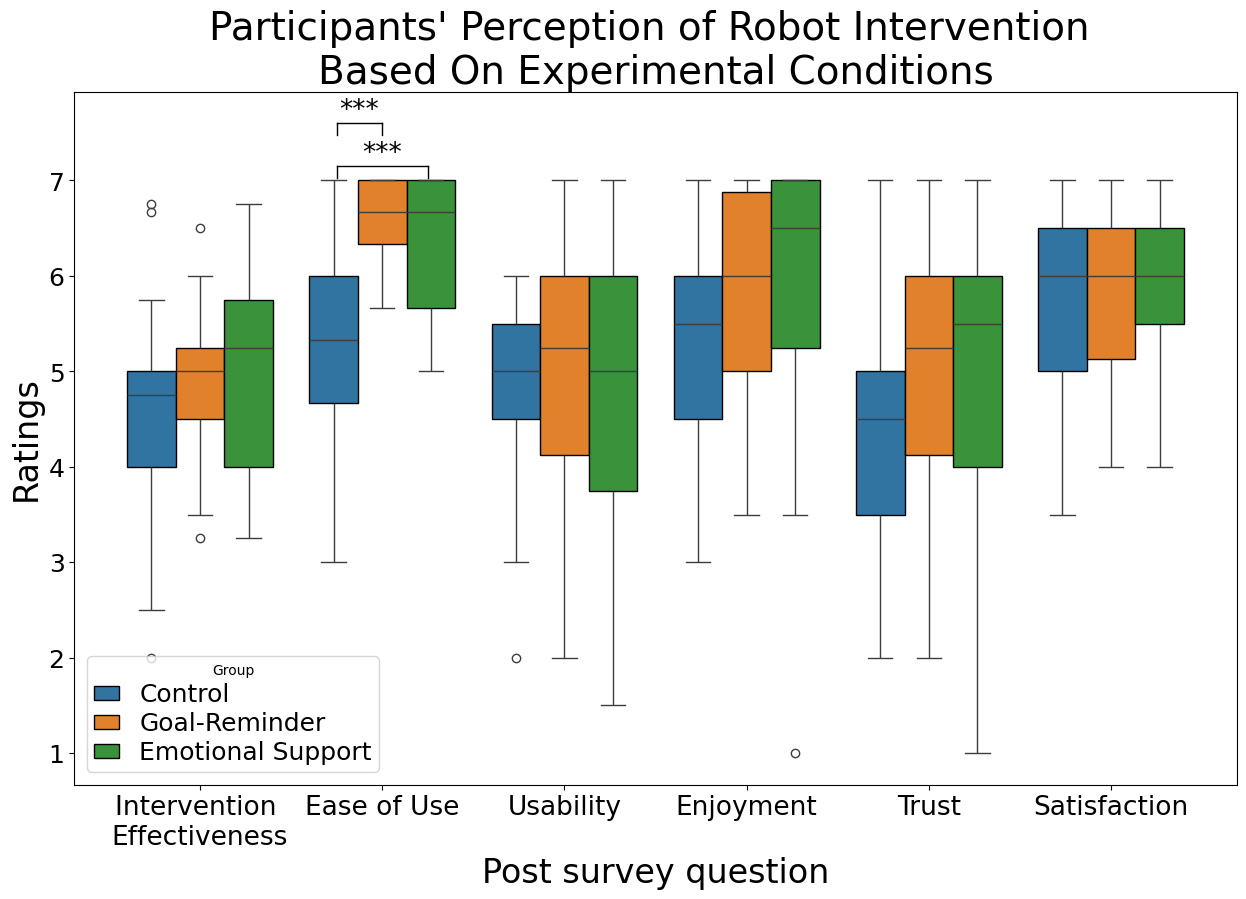}
    \caption{Participants in the \textit{goal reminder} and the \textit{emotional support} conditions reported higher ease of use for our study companion robot than participants in the \textit{control} condition. }
    \vspace{-8pt}
    \label{fig:experimental_conditions+post}
\end{figure}

The Kruskal-Wallis H test showed statistically significant differences in participants' perceived ease of use (PEOU) among the three experimental conditions (\textit{control} $MD$=$5.333$, \textit{goal reminder} $MD$=$6.667$, \textit{emotional support} $MD$=$6.667$; $\chi^2(2)$=$17.502$, $p$=$0.0001$). As illustrated in Fig.~\ref{fig:experimental_conditions+post}, the post-hoc Dunn's test revealed a statistically significant difference between the \textit{control} and the \textit{goal reminder} conditions ($p$=$0.0002$) and the \textit{control} and the \textit{emotional support} conditions ($p$=$0.0008$), but not between the \textit{goal reminder} and the \textit{emotional support} conditions ($p$=$0.604$). 

There was no significant difference in participants' perceived trust among the three experimental conditions (\textit{control} $MD$=$4.5$, \textit{goal reminder} $MD$=$5.25$, \textit{emotional support} $MD$=$5.5$; $\chi^2(2)$=$4.662$, $p$=$0.097$). However, one of the item in the perceived trust sub-scale (``I would rely on the robot for future study sessions.'') showed significant differences (\textit{control} $MD$=$4.0$, \textit{goal reminder} $MD$=$6.0$, \textit{emotional support} $MD$=$5.0$; $\chi^2(2)$=$6.444$, $p$=$0.039$). The post-hoc Dunn's test showed that there was a significant difference between the \textit{control} and the \textit{goal reminder} conditions ($p$=$0.006$), but no significant difference was found between the \textit{control} and the \textit{emotional support} condition ($p$=$0.069$) and between the \textit{goal reminder} and \textit{emotional support} conditions ($p$=$0.308$). We did not find any significant differences in participants' perceived productivity (PP, $\chi^2(2)$=$2.149$, $p$=$0.341$), perceived concentration levels (PC, $\chi^2(2)$=$0.901$, $p$=$0.637$), and effectiveness of robot intervention (ERI, $\chi^2(2)$=$3.504$, $p$=$0.173$).

These results suggest that participants perceived the robot that provides explicit motivational interventions (\textit{goal reminder} and \textit{emotional support} conditions) easier to use than the robot that only provided a physical presence during the self-study task (\textit{control} condition). Also, participants were more willing to use the robot with the \textit{goal-reminder} intervention in the future. 

\raggedbottom

\subsection{Immediate Effect of the Robot Intervention}
We conducted the Mann-Whitney U tests on participants' responses to the picture-based questionnaires (Fig.~\ref{fig:ema}) to compare the effectiveness of the \textit{goal reminder} and the \textit{emotional support} robot interventions. There was no statistically significant difference in the levels of arousal/concentration (goal reminder $MD$=$6.0$, emotional support $MD$=$6.0$, $U$=$5599.0$, $p$=$0.330$); valence/pleasantness (goal reminder $MD$=$5.0$, emotional support $MD$=$5.0$, $U$=$6352.5$, $p$=$0.4967$); and appropriateness of timing (goal reminder $MD$=$6.0$, emotional support $MD$=$6.0$, $U$=$5920.0$, $p$=$0.788$).

\subsection{Effect of the Self-study Goal Achievement}
A Chi-square test of independence was performed to examine the association between the experimental conditions and the level of self-study goal achievement, and we did not find a statistically significant effect ($\chi^2$(4)=$1.182$, $p$=$0.88$). In the \textit{control} condition, there were 5 under-achievers, 7 exact-achievers, and 9 over-achievers. In the \textit{goal reminder} condition, there were 8 under-achievers, 9 exact-achievers, and 9 over-achievers. The \textit{emotional support} condition included 7 under-achievers, 7 exact-achievers, and 13 over-achievers. 

\subsection{Focus Before \& After Each Robot Intervention} 
There were a total of 221 robot interventions delivered between the \textit{goal reminder} and the \textit{emotional support} conditions: $111$ in the \textit{goal reminder} and $110$ in the \textit{emotional support} condition. Of these, seventy-seven interventions were delivered when the participants were distracted, and $79.22$\% (61 out of 77) of these robot interventions were able to have participants re-focus on the self-study task. 
For the \textit{goal reminder} condition, 75.00\% (30 out of 40) of the robot interventions led participants to re-focus on their task. For the \textit{emotional support} condition, 83.78\% (31 out of 37) of the interventions successfully led to re-focus. 

\subsection{Treating the Robot As A Social Other}
Although there was no statistically significant difference in participants' self-study goal achievement or focus behavior change based on the experimental conditions, the behavioral data observed during the study task suggests that participants in the \textit{emotional support} condition treated the robot as a social other more often than participants in the \textit{goal reminder} condition: \textit{control} $MD$=$2.0$; \textit{goal reminder} $MD$=$1.0$; \textit{emotional support} $MD$=$2.5$; $\chi^2(2)$=$6.067$, $p$=$0.048$. The post-hoc Dunn's test showed that \textit{goal reminder} and \textit{emotional support} conditions were significantly different ($p$=$0.048$), while there was no clear difference between \textit{control} and \textit{goal reminder} ($p$=$0.515$) and between \textit{control} and \textit{emotional support} conditions ($p$=$0.161$).

We found similar patterns in the post-study interview data as well. Qualitative analysis on the interview transcripts revealed that each of the robot interventions had different mechanisms in supporting students' focus and productivity during their self-study tasks. Participants who received the \textit{emotional support} intervention highlighted that the robot's encouragement and positive words enhanced their motivation for the study task and tended to refer the robot as a social other rather than a tool. For instance, P15 in the \textit{emotional support} condition said ``I felt more motivated. The robot acted as a {{mentor}}, helping me stay on track and encouraging me throughout the process." P6 also noted ``It kind of felt like having a {{study buddy}}, but a robot.'' and P14 said ``[the robot said] `Oh, we're almost done! Like, we got this!' It's a simple thing, but it definitely helps keeping you motivated to finish whatever you're working on."

On the other hand, participants who received the \textit{goal-reminder} intervention often noted the perceived accountability as the source of the robot's effectiveness and described the robot as a prohibitory tool that prevented them from distractions. For instance, P23 in the \textit{goal reminder} condition noted ``It definitely gave me some peer pressure to focus on my study because it felt like the robot was right in front of me, staring and judging me. That motivated me to finish my work during the time." P12 also shared ``I felt more motivated because of the accountability aspect, like someone else was there with me.''
 
In the \textit{control} condition, the robot had minimal interactions with the participants during the two-hour study sessions, and several participants identified their internal motivation as the main source of their productivity. For instance, P72 in the \textit{control} condition said ``I felt more motivated because I just wanted to finish the work with the time.'' Other participants found the initial SMART goal-setting helpful; ``I felt more motivated because the robot guided me to say the specific short-term goal. So that was really helpful." (P64). However, there were few people who mentioned the mere physical presence of the robot as a study companion.
 
\subsection{Factors Associated with Goal Achievement}
\label{section:task_achivement}
\begin{figure}[t]
    \centering
    \includegraphics[width=0.98\linewidth]{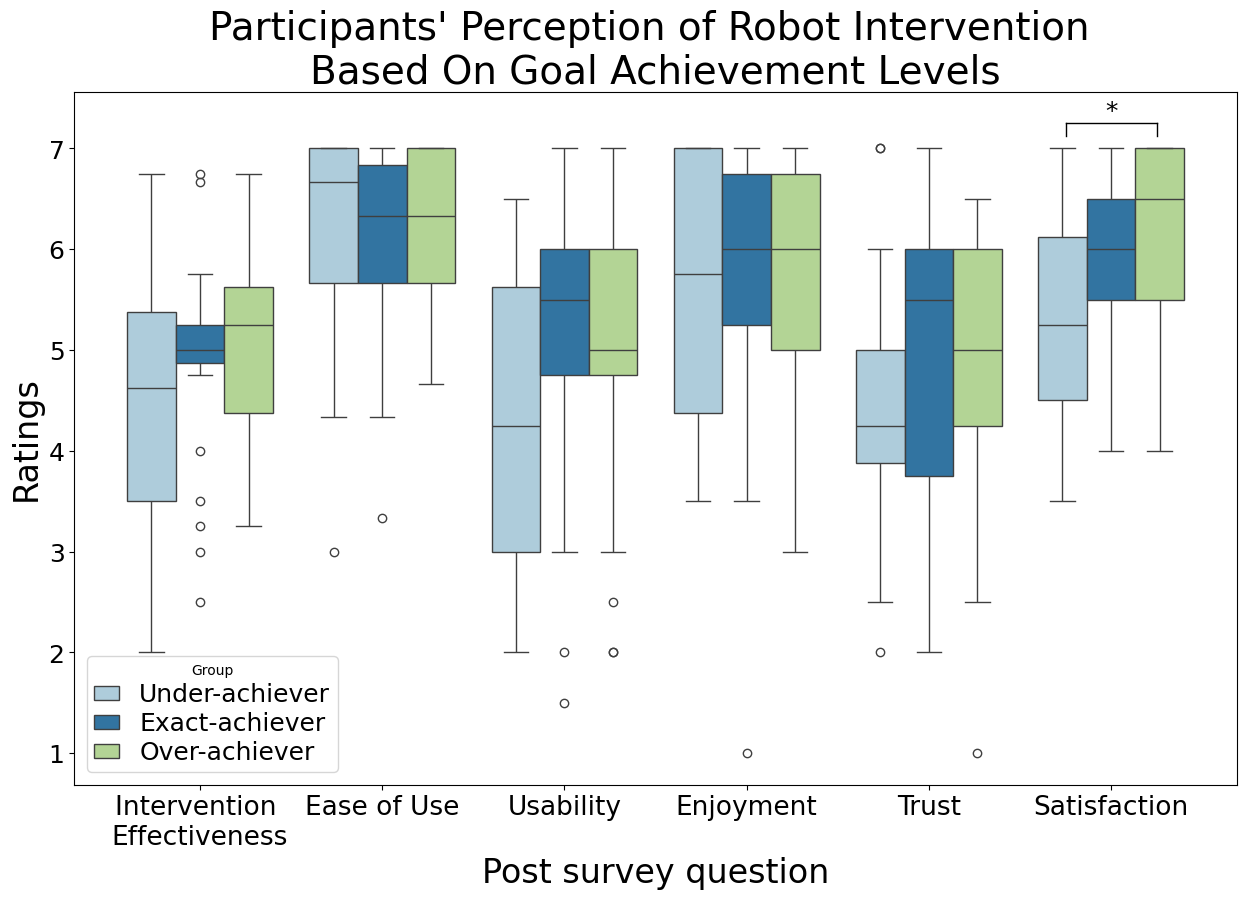}
    \caption{Participants who over-achieved their self-study goals were more satisfied with the robot intervention than participants who under-achieved their goals.}
    \vspace{-8pt}
\label{fig:achievementlevels+experimental_conditions}
\end{figure}
\raggedbottom

We conducted an exploratory analysis to investigate the relationship between participants' SMART goal-achievement and their perception of the robot intervention based on their responses to the post-study questionnaires. There was a statistically significant association between students' goal-achievement level and their satisfaction with the robot intervention: \textit{under-achiever} $MD$=$5.25$, \textit{exact-achiever} $MD$=$6.0$, \textit{over-achiever} $MD$=$6.5$; $\chi^2(2)$=$8.494$, $p$=$0.014$. The post-hoc Dunn's test revealed that the \textit{over-achiever} group's satisfaction was significantly higher than that of the \textit{under-achiever} ($p$=$0.011$) as illustrated in Fig.~\ref{fig:achievementlevels+experimental_conditions}, but there was no difference between \textit{under-achievers} and \textit{exact-achievers} ($p$=$0.095$) and between \textit{exact-achievers} and \textit{over-achievers} ($p$=$0.334$). 
No significant differences in the rest of the items in the post-study questionnaire were found.

We also calculated a Spearman's rank correlation coefficient between participants' satisfaction with the robot and their perception of the robot as a social other, and found that participants who treat the robot as a social agent rather than a device or a tool were more likely to report higher satisfaction with the robot intervention ($\rho(74)$=$0.373$, $p$\textless$0.001$). 

Based on these results, we conducted an ordinal regression analysis to evaluate whether participants' overall satisfaction levels and their perception of the robot as a social other influence significantly predicted their self-study goal achievement levels. The overall regression model was statistically significant with a moderate fit, Cox and Snell's pseudo $R^2$=$0.268$, $\chi^2$(2)=$23.124$, $p$\textless$0.001$. The regression model showed that whether participants perceived the robot as a social other had a significant positive effect on the levels of achievement, $\beta$=$1.903$, $SE$=$0.519$, 95\% Confidence Interval [$0.887$, $2.92$], $p$\textless$0.0001$. The overall satisfaction ratings were not significant in predicting participants' achievement levels ($\beta$=$0.017$, $SE$=$0.012$, 95\% Confidence Interval [$0.006$, $0.040$], $p$=$0.145$). 

\section{Discussion}
Our study explored the efficacy of robotic study companions on participants' focus, motivation, and productivity during self-study sessions. Specifically, we investigated how two different types of robot-mediated motivational interventions that focused on goal-driven support (\textit{goal reminder}) and positive encouragement (\textit{emotional support}) and how their effectiveness were compared to a robot that only provides physical co-presence (\textit{control}) without any explicit support behavior. The \textit{goal reminder} intervention focused on providing repeated reminders about the participants' study goals (instrumental support). On the other hand, the \textit{emotional support} intervention focused on offering emotional support and fostering a sense of companionship for the students in case they were distracted or frustrated with the self-study task. 

In general, participants in all three conditions enjoyed interacting with our study companion robot, as suggested by the overall high satisfaction ratings across the three conditions. Both \textit{goal reminder} and \textit{emotional support} interventions, if delivered when students were distracted, successfully helped them re-focus on their self-study task 75-83\% of the time, and about 70\% of the participants in these conditions either exactly or overly achieved their self-study goals. 

Although \textit{goal reminder} and \textit{emotional support} interventions seem to have similar levels of effectiveness, further behavioral and qualitative analyses suggest that participants who received the \textit{emotional support} intervention were more likely to treat the robot as a social other rather than a productivity tool, and the robot's positive words and encouragement enhanced students' intrinsic motivation and provided a sense of shared experience. 
On the other hand, participants in the \textit{goal reminder} condition described the robot's presence as a ``peer pressure'' that promotes accountability to the goal they have set together and prevents them from distractions.
Lastly, participants in the \textit{control} condition described that the presence of the robot did not affect their motivation, and perceived the robot as a social other the least amount of times compared to both experimental conditions. 

Furthermore, we found that participants' satisfaction with the robot was associated with their level of self-study goal achievement, and their perception of the robot as a social other was a statistically significant predictor of their goal achievement levels. These results suggest that perhaps there is no one single robot-mediated motivational intervention that is superior to others, but it is crucial for the study companion robot to identify the right type of motivational intervention based on each student's contextual needs at the moment and their unique preferences and traits. 

Our robot's \textit{goal reminder} intervention might be appropriate if a student is exhibiting low attention and is spending too much time on tasks outside of their goal. However, the \textit{emotional support} intervention might help build a positive and supportive student-robot relationship for long-term engagement and intervention efficacy.
However, some participants' motivation may have stemmed from being video recorded and observed, rather than solely from the robot’s presence. Additionally, high satisfaction ratings may reflect a novelty effect due to the lab setting and limited exposure. To address this, we plan longer-term studies in more natural environments like students’ homes or libraries to assess sustained impact.
While our study provides initial evidence of the benefits of goal reminding and emotional support in robotic study aids, future long-term studies are needed to investigate whether these positive effects persist and lead to measurable improvements in learning outcomes. 


\section{Conclusion}
In this paper, we present an exploratory study that investigated the opportunities of robotic companions in supporting college students' self-studying. Our study results suggest college students perceived both \textit{goal reminder} and \textit{emotional support} interventions positively, and these interventions successfully helped them re-focus on their study task. However, we found behavioral and qualitative evidence that suggests that the \textit{goal reminder} intervention worked as a regulatory device that prevented students from distractions, and the \textit{emotional support} provided a sense of belonging and shared experience, which led to enhanced intrinsic motivation. 
Our work provides valuable insights to future design and development guidelines for robotic study companions that can effectively support students' motivation and performance in self-study, which could further impact their academic performance and well-being. 

\bibliography{references}

\begin{thebibliography}{10}
\providecommand{\url}[1]{#1}
\csname url@samestyle\endcsname
\providecommand{\newblock}{\relax}
\providecommand{\bibinfo}[2]{#2}
\providecommand{\BIBentrySTDinterwordspacing}{\spaceskip=0pt\relax}
\providecommand{\BIBentryALTinterwordstretchfactor}{4}
\providecommand{\BIBentryALTinterwordspacing}{\spaceskip=\fontdimen2\font plus
\BIBentryALTinterwordstretchfactor\fontdimen3\font minus \fontdimen4\font\relax}
\providecommand{\BIBforeignlanguage}[2]{{%
\expandafter\ifx\csname l@#1\endcsname\relax
\typeout{** WARNING: IEEEtran.bst: No hyphenation pattern has been}%
\typeout{** loaded for the language `#1'. Using the pattern for}%
\typeout{** the default language instead.}%
\else
\language=\csname l@#1\endcsname
\fi
#2}}
\providecommand{\BIBdecl}{\relax}
\BIBdecl

\bibitem{lavy2017benefits}
S.~Lavy, ``Who benefits from group work in higher education? an attachment theory perspective,'' \emph{Higher Education}, vol.~73, pp. 175--187, 2017.

\bibitem{hockings2018independent}
C.~Hockings, L.~Thomas, J.~Ottaway, and R.~Jones, ``Independent learning--what we do when you’re not there,'' \emph{Teaching in Higher Education}, vol.~23, no.~2, pp. 145--161, 2018.

\bibitem{tikkanen2024peer}
L.~Tikkanen, H.~Anttila, S.~Ulmanen, and K.~Pyh{\"a}lt{\"o}, ``Peer relationships and study wellbeing: upper secondary students’ experiences,'' \emph{Social Psychology of Education}, pp. 1--21, 2024.

\bibitem{kosonen2024university}
T.~Kosonen, M.~M{\"a}kinen, J.~Annala, and L.~Penttinen, ``University students’ interpretations of study-related peer sociality,'' \emph{Oxford Review of Education}, vol.~50, no.~5, pp. 710--726, 2024.

\bibitem{byrom2018evaluation}
N.~Byrom, ``An evaluation of a peer support intervention for student mental health,'' \emph{Journal of Mental Health}, vol.~27, no.~3, pp. 240--246, 2018.

\bibitem{altermatt2019academic}
E.~R. Altermatt, ``Academic support from peers as a predictor of academic self-efficacy among college students,'' \emph{Journal of College Student Retention: Research, Theory \& Practice}, vol.~21, no.~1, pp. 21--37, 2019.

\bibitem{cushing1997academic}
L.~S. Cushing and C.~H. Kennedy, ``Academic effects of providing peer support in general education classrooms on students without disabilities,'' \emph{Journal of Applied Behavior Analysis}, vol.~30, no.~1, pp. 139--151, 1997.

\bibitem{milem1998attitude}
J.~F. Milem, ``Attitude change in college students: Examining the effect of college peer groups and faculty normative groups,'' \emph{The Journal of Higher Education}, vol.~69, no.~2, pp. 117--140, 1998.

\bibitem{carter2005effects}
E.~W. Carter, L.~S. Cushing, N.~M. Clark, and C.~H. Kennedy, ``Effects of peer support interventions on students' access to the general curriculum and social interactions,'' \emph{Research and Practice for Persons with Severe Disabilities}, vol.~30, no.~1, pp. 15--25, 2005.

\bibitem{kindermann1993natural}
T.~A. Kindermann, ``Natural peer groups as contexts for individual development: The case of children's motivation in school.'' \emph{Developmental Psychology}, vol.~29, no.~6, p. 970, 1993.

\bibitem{downing2020fear}
V.~R. Downing, K.~M. Cooper, J.~M. Cala, L.~E. Gin, and S.~E. Brownell, ``Fear of negative evaluation and student anxiety in community college active-learning science courses,'' \emph{CBE—life Sciences Education}, vol.~19, no.~2, p. ar20, 2020.

\bibitem{k2021meeting}
M.~K.~Miller, M.~Johannes~Dechant, and R.~L.~Mandryk, ``Meeting you, seeing me: The role of social anxiety, visual feedback, and interface layout in a get-to-know-you task via video chat.'' in \emph{Proceedings of the 2021 CHI Conference on Human Factors in Computing Systems}, 2021, pp. 1--14.

\bibitem{barfield2003students}
R.~L. Barfield, ``Students' perceptions of and satisfaction with group grades and the group experience in the college classroom,'' \emph{Assessment \& Evaluation in Higher Education}, vol.~28, no.~4, pp. 355--370, 2003.

\bibitem{dokuka2015diffusion}
S.~Dokuka, D.~Valeeva, and M.~Yudkevich, ``The diffusion of academic achievements: Social selection and influence in student networks,'' \emph{Higher School of Economics Research Paper No. WP BRP}, vol.~65, 2015.

\bibitem{zimmerman2002becoming}
B.~J. Zimmerman, ``Becoming a self-regulated learner: An overview,'' \emph{Theory into Practice}, vol.~41, no.~2, pp. 64--70, 2002.

\bibitem{shalev2018use}
I.~Shalev, ``Use of a self-regulation failure framework and the nimh research domain criterion (rdoc) to understand the problem of procrastination,'' \emph{Frontiers in Psychiatry}, vol.~9, p. 213, 2018.

\bibitem{muraven2000self}
M.~Muraven and R.~F. Baumeister, ``Self-regulation and depletion of limited resources: Does self-control resemble a muscle?'' \emph{Psychological Bulletin}, vol. 126, no.~2, p. 247, 2000.

\bibitem{park2011teaching}
S.~J. Park, J.~H. Han, B.~H. Kang, and K.~C. Shin, ``Teaching assistant robot, robosem, in english class and practical issues for its diffusion,'' in \emph{Advanced Robotics and its Social Impacts}.\hskip 1em plus 0.5em minus 0.4em\relax IEEE, 2011, pp. 8--11.

\bibitem{dias2008sliding}
M.~B. Dias, B.~Kannan, B.~Browning, E.~G. Jones, B.~Argall, M.~F. Dias, M.~Zinck, M.~M. Veloso, and A.~J. Stentz, ``Sliding autonomy for peer-to-peer human-robot teams,'' in \emph{Intelligent Autonomous Systems 10}.\hskip 1em plus 0.5em minus 0.4em\relax IOS Press, 2008, pp. 332--341.

\bibitem{leite2012modelling}
I.~Leite, A.~Pereira, G.~Castellano, S.~Mascarenhas, C.~Martinho, and A.~Paiva, ``Modelling empathy in social robotic companions,'' in \emph{Advances in User Modeling: UMAP 2011 Workshops, Girona, Spain, July 11-15, 2011, Revised Selected Papers 19}.\hskip 1em plus 0.5em minus 0.4em\relax Springer, 2012, pp. 135--147.

\bibitem{kennedy2015higher}
J.~Kennedy, P.~Baxter, E.~Senft, and T.~Belpaeme, ``Higher nonverbal immediacy leads to greater learning gains in child-robot tutoring interactions,'' in \emph{Social Robotics: 7th International Conference, ICSR 2015, Paris, France, October 26-30, 2015, Proceedings 7}.\hskip 1em plus 0.5em minus 0.4em\relax Springer, 2015, pp. 327--336.

\bibitem{short2014train}
E.~Short, K.~Swift-Spong, J.~Greczek, A.~Ramachandran, A.~Litoiu, E.~C. Grigore, D.~Feil-Seifer, S.~Shuster, J.~J. Lee, S.~Huang \emph{et~al.}, ``How to train your dragonbot: Socially assistive robots for teaching children about nutrition through play,'' in \emph{The 23rd IEEE International Symposium on Robot and Human Interactive Communication}.\hskip 1em plus 0.5em minus 0.4em\relax IEEE, 2014, pp. 924--929.

\bibitem{leyzberg2012physical}
D.~Leyzberg, S.~Spaulding, M.~Toneva, and B.~Scassellati, ``The physical presence of a robot tutor increases cognitive learning gains,'' in \emph{Proceedings of the Annual Meeting of the Cognitive Science Society}, vol.~34, no.~34, 2012.

\bibitem{saerbeck2010expressive}
M.~Saerbeck, T.~Schut, C.~Bartneck, and M.~D. Janse, ``Expressive robots in education: varying the degree of social supportive behavior of a robotic tutor,'' in \emph{Proceedings of the SIGCHI Conference on Human Factors in Computing Systems}, 2010, pp. 1613--1622.

\bibitem{jeong2020robotic}
S.~Jeong, S.~Alghowinem, L.~Aymerich-Franch, K.~Arias, A.~Lapedriza, R.~Picard, H.~W. Park, and C.~Breazeal, ``A robotic positive psychology coach to improve college students’ wellbeing,'' in \emph{2020 29th IEEE International Conference on Robot and Human Interactive Communication (RO-MAN)}.\hskip 1em plus 0.5em minus 0.4em\relax IEEE, 2020, pp. 187--194.

\bibitem{jeong2023deploying}
S.~Jeong, L.~Aymerich-Franch, K.~Arias, S.~Alghowinem, A.~Lapedriza, R.~Picard, H.~W. Park, and C.~Breazeal, ``Deploying a robotic positive psychology coach to improve college students’ psychological well-being,'' \emph{User Modeling and User-Adapted Interaction}, vol.~33, no.~2, pp. 571--615, 2023.

\bibitem{o2024design}
A.~O'Connell, A.~Banga, J.~Ayissi, N.~Yaminrafie, E.~Ko, A.~Le, B.~Cislowski, and M.~Mataric, ``Design and evaluation of a socially assistive robot schoolwork companion for college students with adhd,'' in \emph{Proceedings of the 2024 ACM/IEEE International Conference on Human-Robot Interaction}, 2024, pp. 533--541.

\bibitem{virtanen2015self}
P.~Virtanen, A.~Nevgi, and H.~Niemi, ``Self-regulation in higher education: students’ motivational, regulational and learning strategies, and their relationships to study success,'' \emph{Studies for the Learning Society}, vol.~3, no. 1-2, pp. 20--36, 2015.

\bibitem{dang2023students}
B.~Dang, R.~Vitiello, A.~Nguyen, C.~Ros{\'e}, and S.~J{\"a}rvel{\"a}, ``How do students deliberate for socially shared regulation in collaborative learning? a process-oriented approach,'' in \emph{International Collaboration toward Educational Innovation for All: International Society of the Learning Sciences (ISLS) Annual Meeting 2023}, 2023.

\bibitem{didonato2013effective}
N.~C. DiDonato, ``Effective self-and co-regulation in collaborative learning groups: An analysis of how students regulate problem solving of authentic interdisciplinary tasks,'' \emph{Instructional Science}, vol.~41, pp. 25--47, 2013.

\bibitem{cirillo2018pomodoro}
F.~Cirillo, \emph{The Pomodoro technique: The acclaimed time-management system that has transformed how we work}.\hskip 1em plus 0.5em minus 0.4em\relax Currency, 2018.

\bibitem{pielot2017productive}
M.~Pielot and L.~Rello, ``Productive, anxious, lonely: 24 hours without push notifications,'' in \emph{Proceedings of the 19th International Conference on Human-Computer Interaction with Mobile Devices and Services}, 2017, pp. 1--11.

\bibitem{yang2006exploring}
C.-C. Yang, I.-C. Tsai, B.~Kim, M.-H. Cho, and J.~M. Laffey, ``Exploring the relationships between students' academic motivation and social ability in online learning environments,'' \emph{The Internet and Higher Education}, vol.~9, no.~4, pp. 277--286, 2006.

\bibitem{thoman2007talking}
D.~B. Thoman, C.~Sansone, and M.~Pasupathi, ``Talking about interest: Exploring the role of social interaction for regulating motivation and the interest experience,'' \emph{Journal of Happiness Studies}, vol.~8, pp. 335--370, 2007.

\bibitem{kidd2008robots}
C.~D. Kidd and C.~Breazeal, ``Robots at home: Understanding long-term human-robot interaction,'' in \emph{2008 IEEE/RSJ International Conference on Intelligent Robots and Systems}.\hskip 1em plus 0.5em minus 0.4em\relax IEEE, 2008, pp. 3230--3235.

\bibitem{greczek2014socially}
J.~Greczek, E.~Short, C.~E. Clabaugh, K.~Swift-Spong, and M.~J. Mataric, ``Socially assistive robotics for personalized education for children.'' in \emph{AAAI Fall Symposia}, 2014.

\bibitem{belpaeme2018social}
T.~Belpaeme, J.~Kennedy, A.~Ramachandran, B.~Scassellati, and F.~Tanaka, ``Social robots for education: A review,'' \emph{Science Robotics}, vol.~3, no.~21, p. eaat5954, 2018.

\bibitem{leyzberg2014personalizing}
D.~Leyzberg, S.~Spaulding, and B.~Scassellati, ``Personalizing robot tutors to individuals' learning differences,'' in \emph{Proceedings of the 2014 ACM/IEEE International Conference on Human-Robot Interaction}, 2014, pp. 423--430.

\bibitem{schadenberg2017personalising}
B.~R. Schadenberg, M.~A. Neerincx, F.~Cnossen, and R.~Looije, ``Personalising game difficulty to keep children motivated to play with a social robot: A bayesian approach,'' \emph{Cognitive Systems Research}, vol.~43, pp. 222--231, 2017.

\bibitem{kory2017flat}
J.~M. Kory~Westlund, S.~Jeong, H.~W. Park, S.~Ronfard, A.~Adhikari, P.~L. Harris, D.~DeSteno, and C.~L. Breazeal, ``Flat vs. expressive storytelling: Young children’s learning and retention of a social robot’s narrative,'' \emph{Frontiers in Human Neuroscience}, vol.~11, p. 295, 2017.

\bibitem{ramachandran2017give}
A.~Ramachandran, C.-M. Huang, and B.~Scassellati, ``Give me a break! personalized timing strategies to promote learning in robot-child tutoring,'' in \emph{Proceedings of the 2017 ACM/IEEE International Conference on Human-Robot Interaction}, 2017, pp. 146--155.

\bibitem{gordon2016affective}
G.~Gordon, S.~Spaulding, J.~K. Westlund, J.~J. Lee, L.~Plummer, M.~Martinez, M.~Das, and C.~Breazeal, ``Affective personalization of a social robot tutor for children’s second language skills,'' in \emph{Proceedings of the AAAI Conference on Artificial Intelligence}, vol.~30, no.~1, 2016.

\bibitem{doran1981there}
G.~T. Doran, ``There's a smart way to write managements's goals and objectives.'' \emph{Management Review}, vol.~70, no.~11, 1981.

\bibitem{derryberry2002anxiety}
D.~Derryberry and M.~A. Reed, ``Anxiety-related attentional biases and their regulation by attentional control.'' \emph{Journal of Abnormal Psychology}, vol. 111, no.~2, p. 225, 2002.

\bibitem{attentioncontrolscale}
\BIBentryALTinterwordspacing
{University of Wisconsin–Madison, Affective Neuroscience Laboratory}, ``Attention control scale (attc),'' 2023. [Online]. Available: \url{https://arc.psych.wisc.edu/self-report/attention-control-scale-attc/}
\BIBentrySTDinterwordspacing

\bibitem{heerink2009measuring}
M.~Heerink, B.~Krose, V.~Evers, and B.~Wielinga, ``Measuring acceptance of an assistive social robot: a suggested toolkit,'' in \emph{RO-MAN 2009-The 18th IEEE International Symposium on Robot and Human Interactive Communication}.\hskip 1em plus 0.5em minus 0.4em\relax IEEE, 2009, pp. 528--533.

\bibitem{cox2018analysis}
D.~R. Cox, \emph{Analysis of binary data}.\hskip 1em plus 0.5em minus 0.4em\relax Routledge, 2018.

\bibitem{clarke2017thematic}
V.~Clarke and V.~Braun, ``Thematic analysis,'' \emph{The Journal of Positive psychology}, vol.~12, no.~3, pp. 297--298, 2017.

\end{thebibliography}
\bibliographystyle{IEEEtran}

\end{document}